# A real-time global re-localization framework for 3D LiDAR SLAM

Ziqi Chai, Xiaoyu Shi, Yan Zhou, *Member, IEEE*, and Zhenhua Xiong, *Member, IEEE*

*Abstract*—Simultaneous localization and mapping (*SLAM*) has been a hot research field in the past years. Against the backdrop of more affordable 3D LiDAR sensors, research on 3D LiDAR SLAM is becoming increasingly popular. Furthermore, the re-localization problem with a point cloud map is the foundation for other SLAM applications. In this paper, a template matching framework is proposed to re-localize a robot globally in a 3D LiDAR map. This presents two main challenges. First, most global descriptors for point cloud can only be used for place detection under a small local area. Therefore, in order to re-localize globally in the map, point clouds and descriptors(templates) are densely collected using a reconstructed mesh model at an offline stage by a physical simulation engine to expand the functional distance of point cloud descriptors. Second, the increased number of collected templates makes the matching stage too slow to meet the real-time requirement, for which a cascade matching method is presented for better efficiency. In the experiments, the proposed framework achieves 0.2-meter accuracy at about 10Hz matching speed using pure python implementation with 100k templates, which is effective and efficient for SLAM applications.

## I. INTRODUCTION

SIMULTANEOUS localization and mapping (SLAM) has been an important reseatch topic in the past years. This technology enables automated guided vehicles (AGV) to be applied in various applications, such as disinfection, security inspection, warehousing logistics, smart factory, etc. Furthermore, SLAM-related research has flourished with various sensors, from mono camera to stereo camera, RGBD sensor to light detection and ranging sensor (LiDAR). However, 3D LiDAR SLAM is less developed due to economic constraints than those based on visual sensors and single-line LiDAR. With the development of 3D LiDAR sensors and their affordable price, researchers pay more attention to 3D LiDAR SLAM in recent years.

3D LiDAR, such as Multiline LiDAR, can collect 3D point cloud data in each scan. It usually has a long detecting distance and is robust to light changes, scale-invariant, and suitable for both indoor and outdoor SLAM. With the point cloud data collected at a few positions, a global 3D point cloud map of the environment can be generated. Moreover, one can re-localize its position in the global map with the local scanned data.

To fulfill re-localization in the global 3D point cloud map, some researchers tried to transfer similar ideas from 2D image local invariant descriptors to point clouds. However, a 3D map has a large amount of data, and the processing of 3D data is computationally expensive, which makes it very hard to re-localize in real-time. Other researchers tried to take advantage of the characteristics of point cloud data to realize place recognition by using geometric primitives and global statistics information. However, the research on place recognition mainly focused on Loop Closure. These methods can be used to re-localize the position only near the mapping trajectory and cannot realize re-localization in the whole environment.

In this paper, a real-time matching framework is proposed to re-localize an AGV globally in a pre-built 3D point cloud map. The framework has two stages, namely the offline building stage and the online matching stage. Templates, also called global descriptors, were extracted and organized offline. Then, the AGV can re-localize itself by matching the current scene template in the template library online. Specifically, at the offline building stage, the 3D point cloud map was first built by the SC-LeGO-LOAM algorithm or other dense reconstruction methods. Then, templates are re-collected by a physical simulation engine using a dense reconstructed 3D point cloud map. Finally, the templates are hierarchically clustered. Representative templates of each cluster are used to build a nearest neighbor search engine. At the online matching stage, the Locality Sensitive Hashing (LSH) and K-Dimensional Tree (KD Tree) were used to build a cascade matching pipeline for efficient matching. The localization accuracy was ensured by the offline sampling step, and the online cascade matching stage guaranteed efficiency.

The main features of the proposed re-localization framework can be summarized as follows:
1. It is a Coarse-to-Fine matching framework for global re-localization, not just for Loop Closure.
2. An efficient cascade matching method, consists of LSH and KD Tree, is designed for real-time performance.
3. In the experiments, the proposed framework can achieve a 0.2-meter accuracy at 10Hz using pure python implementation with 100k templates.

The remainder of the paper is organized as follows: in section Ⅱ, related literature is reviewed. Then, the proposed framework is introduced in section Ⅲ, followed by some experimental verifications and discussions in section Ⅳ. Finally, the paper is concluded in section Ⅴ.

This work was supported in part by the National Natural Science Foundation of China (U1813224), Ministry of Education China Mobile Research Fund Project (CMHQ-JS-201900003), and National Key R&D Program of China (2019YFB1310801).

Z.Q. Chai, X.Y. Shi, Y Zhou and Z.H. Xiong are with State Key Laborator y of Mechanical System and Vibration, School of Mechanical Engineering, Sh anghai Jiao Tong University, Shanghai, China (e-mail: chaiziqi@sjtu.edu.cn, 1 5shixiaoyu@sjtu.edu.cn, zhouyan2015@sjtu.edu.cn, mexiong@sjtu.edu.cn).

## II. RELATED WORKS

LIDAR-based place recognition methods can roughly be divided into local descriptors based and global descriptors based methods.

Many local descriptors have been proposed because of their strong matching ability, such as 3D-SURF[1], 3D-SIFT[2], 3D-Harris[3], etc. These descriptors are all extended from their 2D descriptors. PFH[4] and FPFH[5] explored the local surface normal of each neighbor point. SHOT[6] divided the space around a key point into several regions and collects the normal angle histogram of each region to generate the descriptor. Bosse et al. [7] proposed a keypoint voting method. A constant number of nearest neighbor votes per keypoint were queried from a database of local descriptors and aggregated to determine possible place matches. Dubé et al. [8] proposed a method based on the matching of 3D segments. This method extracted segments from a point cloud, matched them with segments from visited places and determines place recognition candidates using a geometric verification step. Guo et al [9] enriched SHOT with the intensity information and proposed a new probabilistic key point voting approach to realize place recognition.

However, the detection of distinctive key points with high repeatability is still a challenging problem. Identifying local descriptors usually requires the extraction of key points and a lot of local geometric calculations.

Compared with local ones, global descriptor matching is more efficient. Wohlkinger et al. [10] proposed ESF, which describes distance, angle, and area distributions on the surface of the partial point cloud using series of histograms. He et al. [11] proposed a novel global point cloud descriptor called M2DP. Instead of analyzing point clouds in three dimensions, the point clouds were projected to many different two-dimensional planes. Then, multiple spatial distribution characteristics of point clouds can be obtained by counting the spatial density histogram of point clouds on the plane. They used the first left singular vector and the right singular vector of these two-dimensional signatures as descriptors. However, it relies on the distribution of all points, and the performance is not satisfied when there is a partial loss of points. Kim et al. [12] proposed an egocentric spatial descriptor named Scan Context. SC encodes a whole point cloud in a 3D LIDAR scan into a matrix using height information of the point cloud. It has shown that extracting only the highest points of a visible point cloud outperforms other existing global descriptors. Based on [12], Wang et al. [13] proposed Intensity Scan Context. They replaced the height information of the point cloud with the intensity and improved the robustness of the descriptor. Compared with [12], Wang et al. [14] encoded the height information to obtain the Lidar-Iris image and applied the Fourier transform on the images to achieve rotation invariance. Jiang et al. [15] formulated a point cloud as a fully connected graph, and the nodes in the graph were used to represent the plane features in the point cloud. This method uses plane geometry constraints to define the similarity between frames. However, many environments do not have rich planar features.

Moreover, all the above methods [11-15] are designed for loop detection, but they do not work well in global place recognition. The main problem is that they can only locate the position near the trajectory of the robot. Cop et al. [16] proposed a point cloud descriptor called DELIGHT. They first divided the 3D space into 16 parts and calculated the histogram of point cloud intensity distribution in each region. A local coordinate system was built to make the descriptor achieve rotational invariance regardless of the viewpoint changes. The disadvantages of this approach are that it is too sensitive to translation and descriptor matching is too time consuming. Some learning-based methods are also proposed. Yin et al[17] proposed a semi-handcrafted deep learning framework called LocNet, which solved the place recognition problem as a similar modeling problem. Uy et al. [18] extracted point cloud descriptors using deep learning, which combines the networks of PointNet [19] and NetVLAD [20]. Kim et al. [21] used CNN to train SC images to realize long-term place recognition. Learning-based methods need a lot of training data and a long training time, and the generalization ability of neural network is limited.

Thus, a real-time global re-localization method is needed. It is not just for Loop Closure, but archives localization globally with real-time performance.

## III. THE PROPOSED RE-LOCALIZATION FRAMEWORK

As shown in Fig.1, the proposed global re-localization framework consists of an offline building stage, including data collection, descriptor extraction, hierarchical clustering, nearest neighbor index building, etc., and an online matching stage, including descriptor extraction, matching with the index, etc.

At the offline stage, a meshed model was reconstructed from the pre-built 3D point cloud map to establish a simulation environment. Then an AGV model is placed at densely sampled positions to collect point cloud data. The Scan Context descriptors in [12] can be extracted from point cloud data. Furthermore, to deal with the rotation invariant issue of the original SC map, a Principal Component Analysis (PCA) was performed on point cloud data first to determine a local reference coordinate while calculating the SC map. Then the extracted descriptor (called the PCA SC map) with metadata (position, PCA angle, each dimension of the PCA SC map, etc.) were finally stored as templates. All templates are clustered, under local connection constraints of their positions, using the hierarchical clustering method to produce representative templates of each cluster, reduce the overall number of templates, and build a hierarchical structure. A nearest neighbor search engine is then built based on these representative templates.

At the online matching stage, the PCA SC map is computed first. Then the normalized column vector of the non-Zero value of each row of the PCA SC map is calculated (called CNZ vector). The CNZ vector is used to search top K candidate presentative templates using the offline-build nearest neighbor search engine. Then a deeper matching will proceed into the clusters for more accurate estimation results.

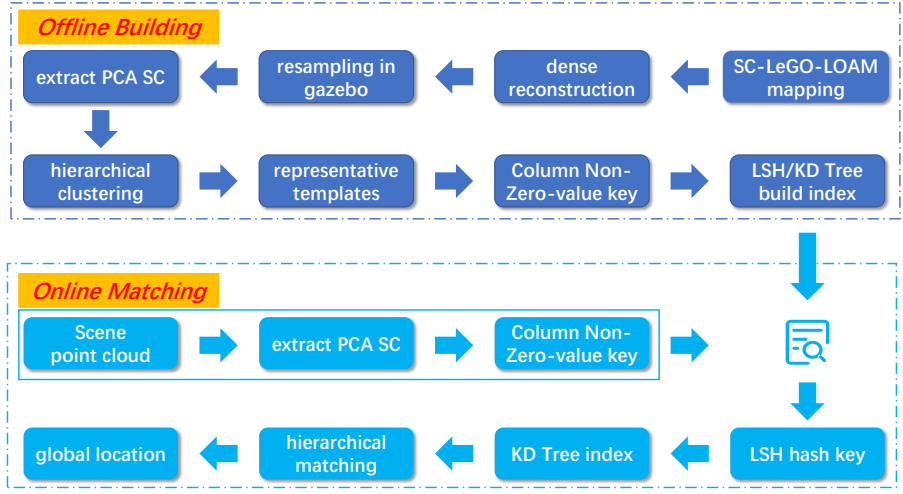

Fig. 1. The proposed pipeline for global re-localization

It is noted that the priority of the proposed framework is to re-localize the AGV under a pre-mapped environment in real-time. It is accomplished by matching scene point cloud descriptors with the descriptor library generated offline. A descriptor is an abstract representation of original data, called templates in this letter, and the re-localization framework is essentially a template matching procedure.

All the symbols used in this section are listed in Table 1.

TABLE I
SYMBOLS USED IN THE PAPER

| Symbol | Meaning |
| --- | --- |
| $d$ | Distance between two samples (two templates) |
| $D$ | Distance between two clusters |
| $s$ | Similarity between two samples (two templates) |
| $s_{max}$ | The max similarity between two samples while shifting the query PCA SC map's column |
| $S$ | Similarity between two clusters |
| $C$ | Cluster |
| $x, y$ | Samples (templates) |
| $m, n$ | A PCA SC map's row and column number |
| $i, j$ | General indices |
| $t_{rep}^{i}$ | representative template of cluster i |

### A. Templates generation

At the present time, the dense 3D reconstruction of unfamiliar environments has become easy using SLAM technologies and products like DJI Terra. In this paper, we adopt the SC-LeGO-LOAM algorithm to mapping the 3D point cloud environment and use Poisson Surface Reconstruction to get the mesh model of the environment. Then a point cloud data re-sampling can be performed using the gazebo simulation engine while moving the AGV in small steps. At each sampling position, the collision between AGV and the environment is detected.

Each frame point cloud data can be cast into a PCA SC descriptor. And a template contains the metadata, such as PCA SC descriptor's height, width, collection position in the map, PCA angle, and the PCA SC descriptor data part. The extracted descriptor (called the PCA SC map here) with metadata were finally stored as templates. The densely sampling procedure in the simulation engine and a visualized example of the PCA SC descriptor can be seen in Fig. 2 and Fig. 3.

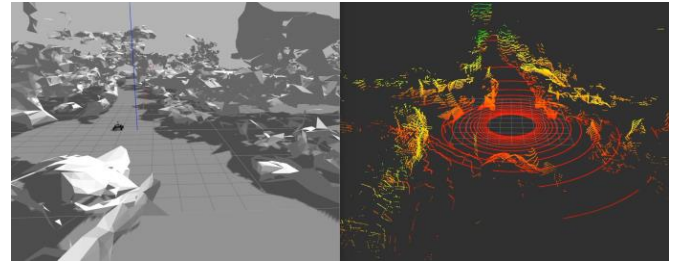

Fig. 2. Re-sampling in gazebo using reconstructed mesh model. Left: AGV in gazebo with mesh model, collecting point cloud data. Right: collected point cloud data.

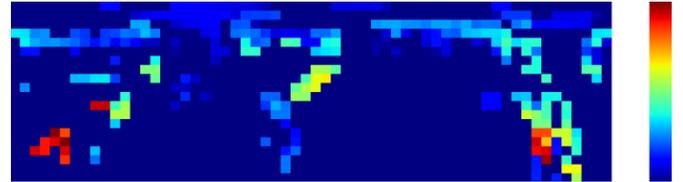

Fig. 3. Extracted PCA SC global descriptor (20 row × 60 column) from point cloud data in Fig. 2.

### B. Template clustering

The oversampled templates are clustered under local connection constraints to produce representative templates of each cluster, reduce the overall number of templates, and build a hierarchical structure.

Templates clustering has two main purposes in this letter. On the one hand, it is a reliable way to generate sparse sampling positions by selecting representative templates in each cluster. The reduced templates can offer a coarse re-localization result very efficiently, and the result can be further refined by combining a point cloud registration. On the other hand, the coarse re-localization result can be refined by deeper matching into the cluster, which is also very efficient.

Templates are clustered from every single sample to high-level clusters, and cluster distances can be measured in many ways. The maximum, minimum, and average linkage matrix is formulated as in (1) (2) and (3).

$$D_{\max}(C_i, C_j) = \max_{x \in C_i, y \in C_j} d(x, y) \quad (1)$$

$$D_{\min}(C_i, C_j) = \min_{x \in C_i, y \in C_j} d(x, y) \quad (2)$$

$$D_{\text{avg}}(C_i, C_j) = \text{avg}_{x \in C_i, y \in C_j} d(x, y) \quad (3)$$

where $C_i$ is the i-th cluster, $D(C_i, C_j)$ is the distance matrix between two clusters to be merged, $x, y$ are samples belonging to each cluster, and $d(x, y)$ is the distance measure between two samples.

As we use an improved Scan Context by PCA, we keep the distances matrix the same way as in [9], e.g. (4).

$$d(x, y) = \frac{1}{n} \sum_{i=1}^{n} \left( \frac{x_i \bullet y_i}{\|x_i\| \bullet \|y_i\|} \right) \quad (4)$$

As the similarity and distance matrix are all normalized from 0 to 1, they can be converted by (5).

$$d(x, y) = 1 - s(x, y) \quad (5)$$

At the building stage, all the templates are built using PCA SC. However, at the matching stage, the templates are matched differently. The PCA SC for the scene is used only when the difference between eigenvalue is larger than a threshold. If so, just a little calculation is needed to compare the two templates. Otherwise, we need to shift the column of SC and calculate the highest score. The similarity matrix at the matching stage is then as in (6) with n equals one or a small range.

$$s_{\max}(x, y) = \max_{i \in n} s(x_i, y) \quad (6)$$

Each cluster's representative templates are selected by maximizing the total similarity between the representative template and all other templates within the cluster as (7).

$$t_{rep}^i = \{x \mid \max_{y \in C_i} \sum s(x, y), x \in C_i\} \quad (7)$$

The clustering procedure is based on similarity calculation between samples. The calculation has a complexity of $O(n^2)$ where $n$ is the total number of samples(templates). On the one hand, the computation cost is unaffordable with $n$ increases. On the other hand, samples far from each other are clustered together, which is not useful to build a local hierarchical structure. So local constraints are introduced while calculating similarities by only considering connected samples and regard others as Zero. Connectivity between samples is determined by searching the sample's top K nearest neighbors (KNN).

*C. Build nearest neighbor search engine*

The number of the presentative templates is much smaller after clustering than that of origin templates. Nevertheless, the total matching time is still growing approximately linearly with the number of templates and the size of the PCA SC map. So, a two-stage cascade matching was proposed to improve the real-time performance. A query vector is first used to determine K candidates in presentative templates using the nearest neighbor search engine, and then the PCA SC map is used for refinement matching in the cluster. This framework's nearest neighbor search engine is built by combining Locality Sensitive Hashing (LSH) and K-Dimensional Trees (KD Tree).

The sparsity of the PCA SC map is used to build the nearest neighbor search engine. We take the normalized non-Zero value of each row as a feature column vector (called CNZ vector). The CNZ vector keeps rotation invariant and compresses the dimension of the data. In practice, the PCA SC map has 20 rows and 60 columns, so we can get a column vector of 20 dimensions with each element normalized by being derived by 60.

The principle of LSH is to calculate a hash-key use feature data, so similar data should produce the same hash-key using a proper hash function. In this way, a subset can be selected just by computing the hash function once. Furthermore, all samples in a subset are stored into a KD Tree to avoid matching query vector exhaustively with samples in the subset. Thus, the LSH can determine one or several subsets in a flash, and the corresponding KD Tree can determine the KNN candidate efficiently.

As for the hash function, we used the binary projection function, which projects the source data vector onto several basis vectors and assigns a binary value (zero or one) according to the sign. Thus, the hash key is the sequence of all binary values (a string like 1101100110). The basis vectors of the hash function are obtained using principal component analysis on the dataset for a more uniform division, called The Principal Component Analysis Binary Projection, for short as PCA-BP. In addition, to make LSH more robust, some other hash functions using random basis vectors are used together, called Random Binary Projection, for short as RBP.

At the online matching stage, the PCA SC and its CNZ vector are computed. First, the CNZ vector is used to search top K candidate presentative templates using the offline-build engine. Then a deeper matching can proceed into the clusters for accurate location estimation.

IV. EXPERIMENTAL RESULTS

In this section, experiments were mainly carried on simulation data because the performance of the proposed re-localization framework is independent of how these data are collected. During the following experiments, 100K templates are generated at the offline stage. All the time metrics are equivalent to single-threaded Python implementation.

In subsection A, we show the visualization results of hierarchical clustering and illustrate the importance of local constraints during similarity calculation in terms of saving total clustering time while not decrease clustering performance. In subsection B, we demonstrate the feasibility of enabling global descriptors re-localizing globally in 3D point cloud map by using the offline dense sampling method. Results in this subsection are obtained without any acceleration, and the only purpose is to show the globally re-localization ability. At last, we prove the real-time positioning capability of the cascading matching framework in subsection C. Also, the migration potential and predictable performance of the framework for other global descriptors are discussed.

*A. Validation of clustering*

In this subsection, first, we compared the performance of different clustering principles on the same data. The visualization results of clustering under different principles are shown in Fig. 4. For better visualization, only 10K samples were used.

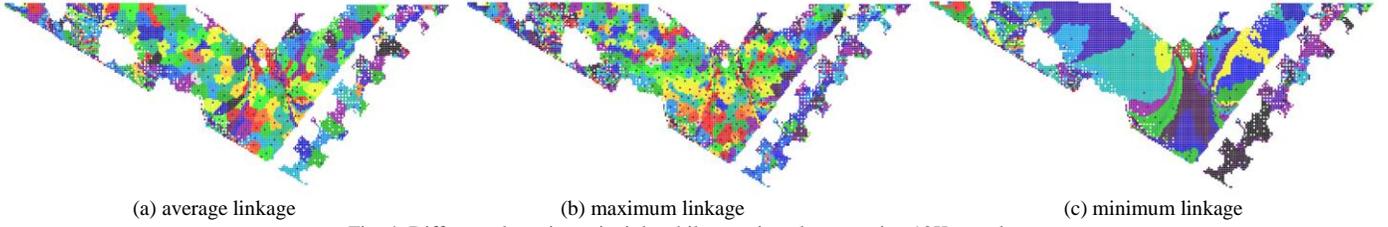

(a) average linkage  (b) maximum linkage  (c) minimum linkage

Fig. 4. Different clustering principle while merging clusters using 10K samples.

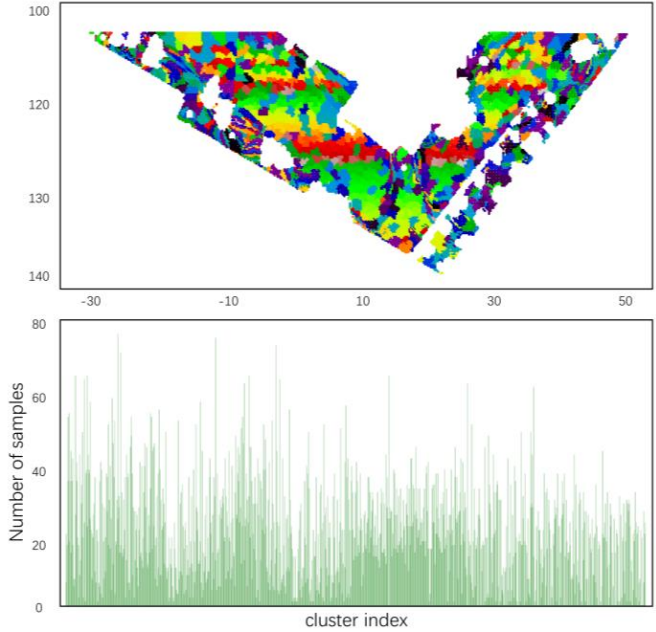

Fig. 5. Clustered results using maximum linkage on 10K samples using KNN local constrains. The cluster threshold is 0.4 in this figure. Left: visualization of clusters identified by colors. Right: number of samples in each cluster.

Representative templates of each cluster

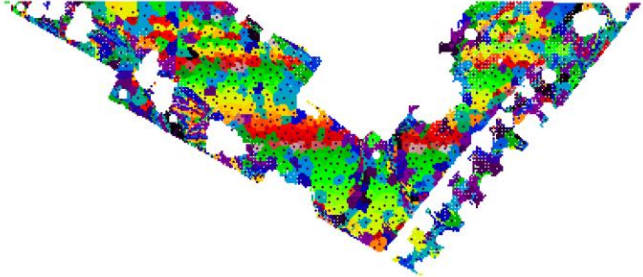

Fig. 6. Representative templates of each cluster are plotted with black spots.

As we can see, maximum linkage tries to constrain the maximum distance between each pair of samples in the cluster and produce the best local constraints. In contrast, Minimum linkage tries to constrain the minimum distance between each pair of samples in the cluster and prefers to link more clusters into one, which produces the worst cluster result. When using maximum linkage as cluster principle, KNN as local constrains, 0.4 as cluster threshold, then the average number of each cluster is above 20 as shown in Fig. 5. And representative templates are plotted by black spots in Fig. 6.

Second, to prove that local constraints during the clustering procedure can distinctively decrease the total clustering time while not decrease the matching accuracy in the meantime, we repeat clustering and matching while with and without local constraints. The statistical results in Table 2 were collected using 100K samples, clustered by the maximum linkage principle. In practice, all computations use 96 threads, but all the time indicators shown in Table 2 are converted to the single-threaded equivalent time.

The clustered results in Table 2 are similar in the following aspects: the number of clusters, time for clustering, and re-localization errors. It shows the local connection constraints have no negative influence on clustering performance. In contrast, the time cost by similarity calculation when use None local constrain is 2000 times that of using KNN constrain. As observed in the table, the number of clusters without constraints is a little smaller, and the success rate is a little higher because similar samples are clustered together despite how far they are apart from each other. All things considered, using local constraints during similarity calculation is important in terms of saving total clustering time while not decrease clustering performance.

### B. Validation of global re-localization

In order to show the global re-localize ability of the proposed framework, the PCA SC maps of trajectory points and the PCA SC maps of re-sampled points are used to carrying out a cross-matching in this subsection. During testing, all the 100K samples are divided into templates samples(50K) and test samples(50K) when using re-sampled templates to match re-sampled templates scenes. Results in this subsection are obtained without any acceleration. The re-localization is accomplished by only exhaustively matching because the only purpose is to show the globally re-localization ability.

The statistical results in Table 3 show that the PCA SC collected during mapping using SC-LeGO LOAM can only be used for Loop Closure. Furthermore, when tested by the PCA SC of re-sampled points, the success rate is extremely low and success only when the test point is within a limited distance to the trajectory path. In contrast, the PCA SC of re-sampled points can deal with all test cases, including trajectory points.

The results indicate that global descriptors like SC map have a limitation when used for global re-localization because of the lack of translation invariant. And it is feasible to re-localize after densely offline sampling.

### C. Validation of real-time re-localization

The proposed matching framework is a real-time re-localization system. In this subsection, we tested the real-time performance of the framework using 100K templates.

In the test pipeline, the PCA SC was first extracted from the test sample and produced a CNZ vector, which was then used as the nearest neighbor search key. The nearest neighbor search engine is built as a cascade structure of LSH and KD Tree using CNZ vectors of all representative templates of all clusters as keys. The nearest neighbor search engine is used to determine top candidates among representative templates of all clusters. After top KNN candidates were determined, a deeper search is performed into clusters for a more accurate re-localization result. There is a tradeoff between speed and precision of matching. They are closely related to the number of KNN candidates.

The performance of the pure KD Tree is as shown in Table 4. At the online matching stage, only KD Tree is used for candidates searching among representative templates. The cluster threshold indicates the maximum similarity distance among all templates within one cluster. When the cluster threshold is set to 0.1, the number of KNN candidates is set to 10, the average time for matching is about 25ms, 40Hz, with a success rate of 97%, when the re-localization error is less than 0.2 meters. Considering the time for calculating one PCA SC descriptor is about 60ms, then the re-localization speed is about 10Hz overall.

TABLE II
HIERARCHICAL CLUSTERING WITH OR WITHOUT LOCAL CONSTRAINS

| Number of samples | Local constrain | Similarity compute time (s) | Cluster threshold | Cluster time (s) | Number of clusters | Match time (s) | Percentage within re-localization error | | | | |
|---|---|---|---|---|---|---|---|---|---|---|---|
| | | | | | | | 0.2m | 0.4m | 0.6m | 0.8m | 1.0m |
| 100K | None | 228.096 | 0.2 | 669.78 | 15162 | 15.55 | 0.99999 | 1 | 1 | 1 | 1 |
| | | | 0.3 | 600.18 | 7708 | 8.00 | 0.99994 | 0.99998 | 0.99998 | 0.99998 | 0.99998 |
| | | | 0.4 | 600.78 | 4375 | 4.69 | 0.99934 | 0.99954 | 0.99966 | 0.99972 | 0.99976 |
| | | | 0.5 | 589.49 | 2447 | 2.90 | 0.99723 | 0.99795 | 0.99848 | 0.9988 | 0.99895 |
| | KNN | 0.108 | 0.2 | 386.15 | 15440 | 16.14 | 0.99996 | 1 | 1 | 1 | 1 |
| | | | 0.3 | 390.49 | 8352 | 8.74 | 0.99991 | 0.99998 | 0.99998 | 0.99998 | 0.99998 |
| | | | 0.4 | 365.81 | 5906 | 6.23 | 0.99771 | 0.99884 | 0.9992 | 0.99935 | 0.9994 |
| | | | 0.5 | 366.31 | 4579 | 4.91 | 0.98802 | 0.99304 | 0.99612 | 0.99698 | 0.99749 |

Percentage within re-localization error means the percentage that how many test-samples' re-localization error is smaller than the threshold. The similarity calculation time and the matching time are the average time for one scene using python implementation with one thread.

TABLE III
ABILITY TO RE-LOCALIZE GLOBALLY(WITHOUT ANY ACCELERATION)

| Scene | Number of scenes | Template | Number of templates | Match time (s) | Percentage within re-localization error | | | | |
|---|---|---|---|---|---|---|---|---|---|
| | | | | | 0.2m | 0.4m | 0.6m | 0.8m | 1.0m |
| Re-sampled points | 50K | Re-sampled points | 50K | 66.290 | 0.3970 | 0.9615 | 0.9819 | 0.9872 | 0.9905 |
| Re-sampled points | 100K | Trajectory points | 1.6K | 2.295 | 0.0239 | 0.0502 | 0.0778 | 0.1007 | 0.1251 |
| Trajectory points | 1.6K | Re-sampled points | 100K | 139.249 | 1 | 1 | 1 | 1 | 1 |

Scene PCA SC are matched with Templates PCA SC without any acceleration. The *Trajectory points* means the PCA SC are collected using point cloud data during mapping, while the *Re-sampled points* are generated offline. The matching time is the average time for matching one scene using python implementation.

TABLE IV
REAL-TIME PERFORMANCE OF KD TREE

| Number of samples | Cluster threshold | Number of KNN | Search KNN time (ms) | Match cluster time (ms) | Percentage within re-localization error | | | | |
|---|---|---|---|---|---|---|---|---|---|
| | | | | | 0.2m | 0.4m | 0.6m | 0.8m | 1.0m |
| 100K | 0.1 | 10 | 0.9121 | 24.8203 | 0.98143 | 0.99649 | 0.99849 | 0.99895 | 0.99930 |
| | 0.2 | 10 | 0.6230 | 71.4284 | 0.97319 | 0.98936 | 0.99424 | 0.99595 | 0.99706 |
| | 0.3 | 10 | 0.5024 | 138.3669 | 0.96060 | 0.97914 | 0.98733 | 0.99053 | 0.99268 |
| | 0.4 | 10 | 0.4572 | 177.0508 | 0.95068 | 0.97181 | 0.98193 | 0.98572 | 0.98858 |
| | 0.5 | 10 | 0.4568 | 209.6878 | 0.94821 | 0.96750 | 0.97769 | 0.98198 | 0.98500 |
| | 0.6 | 10 | 0.4483 | 228.5109 | 0.94444 | 0.96335 | 0.97398 | 0.97841 | 0.98148 |

The data shown in this table was collected using 100K samples, clustered by complete linkage principle. The unit for time is millisecond, the unit for re-localization error is meter. Percentage within re-localization error means the percentage that how many test-samples' re-localization error is smaller than the threshold. The matching time is the average time for matching one scene using python implementation.

TABLE V
REAL-TIME PERFORMANCE WITH LSH AND KD TREE

| Number of samples | Cluster threshold | Number of KNN | Search KNN time (ms) | Match cluster time (ms) | Percentage within re-localization error | | | | |
|---|---|---|---|---|---|---|---|---|---|
| | | | | | 0.2m | 0.4m | 0.6m | 0.8m | 1.0m |
| 100K | 0.1 | 10 | 1.1203 | 25.5830 | 0.97588 | 0.99514 | 0.99799 | 0.99873 | 0.99915 |
| | 0.2 | 10 | 0.7345 | 70.0790 | 0.96103 | 0.98375 | 0.99152 | 0.99428 | 0.99603 |
| | 0.3 | 10 | 0.5411 | 136.2638 | 0.94319 | 0.96869 | 0.98115 | 0.98636 | 0.98970 |
| | 0.4 | 10 | 0.4860 | 173.4381 | 0.93221 | 0.95958 | 0.97367 | 0.97991 | 0.98422 |
| | 0.5 | 10 | 0.4362 | 206.9575 | 0.92579 | 0.95205 | 0.96756 | 0.97452 | 0.97926 |
| | 0.6 | 10 | 0.4089 | 223.9088 | 0.92241 | 0.94785 | 0.96354 | 0.97040 | 0.97540 |

The data shown in this table was collected using 100K samples, clustered by complete linkage principle. The unit for time is millisecond, the unit for re-localization error is meter. Percentage within re-localization error means the percentage that how many test-samples' re-localization error is smaller than the threshold. The matching time is the average time for matching one scene using python implementation.

The performance of a combination of LSH and KD Tree is as shown in Table 5. At the online matching stage, the nearest neighbor searching engine is the combination of LSH and KD Tree. In Table 5, both the efficiency and accuracy are similar to that of the pure KD Tree situation. This is because many optimizations have been applied to the pure KD Tree method in this paper, such as using a CNZ feature vector to replace the PCA SC feature vector to reduce feature dimension and cluster templates to reduce the total number of templates samples. So, the advantage of the combination of LSH and KD Tree is more obvious when expanding the number of samples and using high-dimensional features. The searching time varies with the number of samples and the dimension of each sample, as shown in Fig. 7 for the pure KD Tree method and as in Fig. 8 for the combined LSH-KD Tree method. Fig.7 and Fig. 8 show the potential to use the framework on large-scale mapping and transfer it to other descriptors.

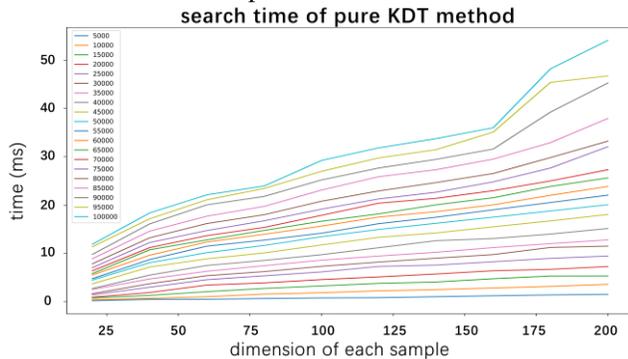

Fig. 7. Searching time for pure KD Tree is increasing with the total number of samples and the dimensions of each sample. Number of total count of samples are identified by colors.

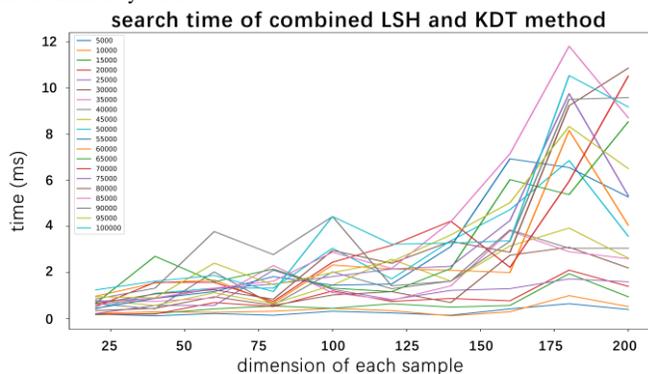

Fig. 8. Searching time for combined LSH and KD Tree increase slower with the total count of samples and the dimensions of each sample. Number of total count of samples are identified by colors.

## V. CONCLUSION

With the development of 3D LiDAR sensors, SLAM using 3D LiDAR has been popular these years. A cascade template matching framework was proposed in this paper to re-localize in 3D point cloud map in real-time. First, point cloud data were densely sampled in simulation using a dense reconstructed mesh model, followed by feature extraction. Then a framework that consists of clustering and nearest neighbor searching was built to match templates online.

When testing with pure python implementation on one thread, the framework achieved 10Hz matching 100K templates. Ignoring the time for calculating the PCA SC descriptor, the matching speed is about 40Hz. The matching speed is considerable when the framework is set up on a cloud server offering re-localization services. Moreover, the successful match rate is 97% when the re-localize error is less than 0.2m, 99% when the re-localize error is less than 0.4m. Moreover, the efficiency of the framework can be greatly improved when optimized using the C++ language, SIMD instruction, multiprocessing coding, etc., which is predictable.

REFERENCES

[1] KNOPP J, PRASAD M, WILLEMS G, et al. Hough transform and 3D SURF for robust three dimensional classification; proceedings of the European Conference on Computer Vision, F, 2010 [C]. Springer.
[2] SCOVANNER P, ALI S, SHAH M. A 3-dimensional sift descriptor and its application to action recognition; proceedings of the Proceedings of the 15th ACM international conference on Multimedia, F, 2007 [C].
[3] SIPIRAN I, BUSTOS B. A Robust 3D Interest Points Detector Based on Harris Operator [J]. 3DOR, 2010, 2(3.
[4] RUSU R B, BLODOW N, MARTON Z C, et al. Aligning point cloud views using persistent feature histograms; proceedings of the 2008 IEEE/RSJ international conference on intelligent robots and systems, F, 2008 [C]. IEEE.
[5] RUSU R B, BLODOW N, BEETZ M. Fast Point Feature Histograms (FPFH) for 3D registration; proceedings of the 2009 IEEE International Conference on Robotics and Automation, F 12-17 May 2009, 2009 [C].
[6] SALTI S, TOMBARI F, DI STEFANO L. SHOT: Unique signatures of histograms for surface and texture description [J]. Computer Vision and Image Understanding, 2014, 125(251-64.
[7] BOSSE M, ZLOT R. Place recognition using keypoint voting in large 3D lidar datasets; proceedings of the 2013 IEEE International Conference on Robotics and Automation, F 6-10 May 2013, 2013 [C].
[8] DUBé R, DUGAS D, STUMM E, et al. SegMatch: Segment based place recognition in 3D point clouds; proceedings of the 2017 IEEE International Conference on Robotics and Automation (ICRA), F 29 May-3 June 2017, 2017 [C].
[9] GUO J D, BORGES P V K, PARK C, et al. Local Descriptor for Robust Place Recognition Using LiDAR Intensity [J]. Ieee Robotics and Automation Letters, 2019, 4(2): 1470-7.
[10] WOHLKINGER W, VINCZE M. Ensemble of shape functions for 3D object classification; proceedings of the 2011 IEEE International Conference on Robotics and Biomimetics, F 7-11 Dec. 2011, 2011 [C].
[11] HE L, WANG X, ZHANG H. M2DP: A novel 3D point cloud descriptor and its application in loop closure detection; proceedings of the 2016 IEEE/RSJ International Conference on Intelligent Robots and Systems (IROS), F 9-14 Oct. 2016, 2016 [C].
[12] KIM G, KIM A. Scan Context: Egocentric Spatial Descriptor for Place Recognition Within 3D Point Cloud Map; proceedings of the 2018 IEEE/RSJ International Conference on Intelligent Robots and Systems (IROS), F 1-5 Oct. 2018, 2018 [C].
[13] WANG H, WANG C, XIE L. Intensity Scan Context: Coding Intensity and Geometry Relations for Loop Closure Detection; proceedings of the 2020 IEEE International Conference on Robotics and Automation (ICRA), F 31 May-31 Aug. 2020, 2020 [C].
[14] WANG Y, SUN Z, XU C-Z, et al. LiDAR Iris for Loop-Closure Detection [J]. arXiv preprint arXiv:191203825,
[15] JIANG J W, WANG J K, WANG P, et al. LiPMatch: LiDAR Point Cloud Plane Based Loop-Closure [J]. Ieee Robotics and Automation Letters, 2020, 5(4): 6861-8.
[16] COP K P, BORGES P V K, DUBé R. Delight: An Efficient Descriptor for Global Localisation Using LiDAR Intensities; proceedings of the 2018 IEEE International Conference on Robotics and Automation (ICRA), F 21-25 May 2018, 2018 [C].
[17] YIN H, TANG L, DING X, et al. LocNet: Global Localization in 3D Point Clouds for Mobile Vehicles; proceedings of the 2018 IEEE Intelligent Vehicles Symposium (IV), F 26-30 June 2018, 2018 [C].
[18] UY M A, LEE G H. PointNetVLAD: Deep Point Cloud Based Retrieval for Large-Scale Place Recognition; proceedings of the 2018 IEEE/CVF Conference on Computer Vision and Pattern Recognition (CVPR), F 2018/june]. IEEE Computer Society.
[19] QI C R, SU H, MO K, et al. PointNet: Deep Learning on Point Sets for 3D Classification and Segmentation; proceedings of the Proceedings of the IEEE Conference on Computer Vision and Pattern Recognition (CVPR), F 2017/july].
[20] ARANDJELOVIĆ R, GRONAT P, TORII A, et al. NetVLAD: CNN Architecture for Weakly Supervised Place Recognition [J]. IEEE Transactions on Pattern Analysis and Machine Intelligence, 2018, 40(6): 1437-51.
[21] KIM G, PARK B, KIM A. 1-Day Learning, 1-Year Localization: Long-Term LiDAR Localization Using Scan Context Image [J]. Ieee Robotics and Automation Letters, 2019, 4(2): 1948-55.